\documentclass[runningheads]{llncs}
\usepackage{hyperref}
\usepackage{graphicx}
\usepackage{cite}
\usepackage{xurl}
\usepackage{mathtools,amssymb}
\usepackage{amsmath}
\usepackage{threeparttable}
\begin{document}

\title{SOMPS-Net : Attention based social graph framework for early detection of fake health news}
%
%

\author{Prasannakumaran D\inst{1} \and
Harish Srinivasan\inst{2} \and
Sowmiya Sree S \inst{3}\and
Sri Gayathri Devi I \inst{3} \and
Saikrishnan S \inst{1} \and
Vineeth Vijayaraghavan \inst{2}}
\authorrunning{Prasannakumaran et al.}
%
\institute{SSN College of Engineering, Chennai, India  \\   
\email{\{prasannakumaran18110,saikrishnan18133\}@cse.ssn.edu.in} \and
Solarillion Foundation, Chennai, India\\
\email{harishsrinivasan1999@gmail.com}\\
\email{vineethv@ieee.org} \and 
College of Engineering Guindy, Anna University, Chennai, India \\
\email{\{soumiya2805,gayathri2800\}@gmail.com}}
\titlerunning{SOMPS-Net: Attention based social graph framework}

\maketitle
\begin{abstract}
Fake news is fabricated information that is presented as genuine, with intention to deceive the reader. Recently, the magnitude of people relying on social media for news consumption has increased significantly. Owing to this rapid increase, the adverse effects of misinformation affect a wider audience. 
On account of the increased vulnerability of people to such deceptive fake news, a reliable technique to detect misinformation at its early stages is imperative. Hence, the authors propose a novel graph-based framework \textbf{SO}cial graph with \textbf{M}ulti-head attention and \textbf{P}ublisher information and news \textbf{S}tatistics \textbf{Net}work (SOMPS-Net) comprising of two components -- \textit{Social Interaction Graph} (SIG) and \textit{Publisher and News Statistics} (PNS). The posited model is experimented on the HealthStory dataset and generalizes across diverse medical topics including Cancer, Alzheimer's, Obstetrics, and Nutrition. SOMPS-Net significantly outperformed other state-of-the-art graph-based models experimented on HealthStory by 17.1\%. Further, experiments on early detection demonstrated that SOMPS-Net predicted fake news articles with 79\% certainty within just 8 hours of its broadcast. Thus the contributions of this work lay down the foundation for capturing fake health news across multiple medical topics at its early stages.

\end{abstract}
\section{Introduction} \label{introduction}


The onset of digitization has deemed social media to be a major source for news consumption. This has also resulted in the widespread diffusion of misinformation, widely known as \textit{fake news}. \cite{study1} revealed that an agency operated dozens of Twitter accounts masquerading as local news sources that collectively garnered more than half-a-million followers. One of the major reasons for this prevalent dissemination is stated by the \textit{Social Identity Theory} \cite{socialtheory}; people belonging to the same group favor the social group they are involved in. This perceived credibility hinders an individual from validating any or some of the news consumed from that group. Furthermore, the impacts of social media are not limited to politics, it also poses a huge threat to global health. A study \cite{usAdults_study} showed that 60\% of US adults consumed health information from social media.
The problem cascades further when misinformation is spread by credible sources.

Considering these adverse effects of misinformation, extensive research work have been carried out to tackle fake news spread. According to \cite{fnd_approaches}, as of 2020, majority of the existing approaches used content based models. They use the textual and visual information available in the news article. However, the methods used for fabricating fake news is evolving such that they are indistinguishable from real news. This calls for additional reliable information to be jointly explored with news content for improved fake news detection. 

Social media has become the main source of news online with more than 2.4 billion internet users. Hence, Shu et al. \cite{userDynamics} studied the user dynamics and \cite{userProfile} utilised the attributed information of users that are extracted from their accounts. The profile features of users such as whether the account is verified or not, number of followers and user-defined location are some of the potential factors to differentiate authentic and malicious users. Social media platforms like Twitter are used not just by millions of humans but also by innumerable bots which are designed to mimic human behaviour. According to an estimate in 2017 \cite{social_bots}, there were 23 million bots on Twitter which comprises 8.5\% of all accounts. Bots create bursts of tweets about an issue and \cite{botsatearlystage} claims that they are particularly active in the early spreading phases of viral claims. They inflate the popularity of fake news and hence contribute remarkably to its spread.
Also, individual user Twitter engagements (tweets/retweets/replies) collectively help to perceive a community wide opinion about a given news article. 

With increase in sensitivity of news, the social engagements associated with it increases exponentially. Such multitude of social engagements and user connections are best captured and visually represented using graphs. Graph Neural Networks (GNNs) are a class of deep learning methods designed to infer data described by graphs. GNNs have the capability to fuse heterogeneous data like engagements posted by an user and their profile activity. Hence, the inherent relational and logical information about a given news can be captured using GNN.

The main contributions of this research work are summarised as follows.
\begin{enumerate}
\item A novel graph-based framework that jointly utilizes social engagements of associated users along with the publisher details and social media statistics of the news article. 

\item With extensive use of robust user and news features, the posited model accurately predicts the 
veracity of a news article at early stages to combat its spread.

\item The authors demonstrate the superiority of the proposed model over established baselines that consider health articles across various subjects. 
\end{enumerate}

The course of the research work is organized as follows. \hyperref[related work]{Section 2} discusses relevant research in fake news detection. The details of the dataset used in this work are given in \hyperref[dataset]{Section 3}. The problem definition is presented in \hyperref[problem defn]{Section 4}. \hyperref[architecture]{Section 5} elucidates the component in the proposed architecture. The results of the work and other experiments are illustrated in \hyperref[results]{Section 6}. Finally the authors conclude the work and discuss future scope of this research in \hyperref[conclusion]{Section 7}.

\section{Related Work}
\label{related work}

Fake news detection has been an active research area in the recent past. Several solutions have been proposed to detect fake news and combat its spread. 
Wynne et al. \cite{Wynne} utilized vocabulary of the news article and \cite{sentiment,rubin-etal-2016-fake} used the linguistic features to detect fake news. Furthermore, the visual content present in articles has also been utilized to tackle the problem. 
In the real world, fake-news image differs from true-news images at both physical and semantic levels. The image quality and the characteristics of the pixel attribute to the distinction between these images at the physical and semantic level respectively.
At the physical level, fake-news images might be of low quality, which is reflected in the frequency domain. At the semantic level, the images exhibit distinct characteristics in the pixel domain. 
Hence, \cite{MVNN} captured complex patterns of fake news images in the frequency domain and extracted visual features from different semantic levels in the pixel domain for fake news detection.

Zhou et al. \cite{SAFE} studied the cross modal relationships between text and visual information and concluded that the two complement each other and therefore were utilized together for detecting fake news articles. However, due to limited fact-checking experts to verify a news article, improved writing style of fake news spreaders and advancements in image manipulation techniques, the aforementioned solutions are suboptimal.

An alternative approach would be to extract details about the news from a reliable platform to evaluate its authenticity. It is well established that almost every news reaches the public via social media platforms. To strengthen this claim, the fake news triangle proposed in \cite{fakenewstriangle}, posited three items (Social network, motivation, tools \& services) and claimed that without any one of these factors, fake news diffuses at a slower rate. 
Hence, online social media platforms were asserted as inevitable tools for the spread of fake news. 
Therefore, the diversified information available in social media provides multiperspectivity of the news to aid the detection task. Liu et al. \cite{Liu} exploited user profile features from social media platforms such as Twitter and Weibo. Studies \cite{fakenewsfast, fakeearlyspread} have asserted that fake news spreads much faster than true news and also investigated the dynamic evolution of propagation topology. In support to this assertion, Wu et al. \cite{traceminer} used the propagation path of a news article to detect the veracity of a Twitter post. 

As stated in Section \ref{introduction}, the necessity and prospects of graphs to represent data from social media has motivated several research groups \cite{k-graph, GCAL, gcan, SCARLET} to exploit GNNs for the detection task. In \cite{k-graph} positive and negative knowledge graphs were constructed for detecting fake news. Further, \cite{GCAL} claimed that integrating user's comment and content of the article in heterogeneous graph improved the detection rate. Lu et al. \cite{gcan} proposed an explainable graph model for fake news detection and Rath et al. \cite{SCARLET} utilised trust and credibility scores of users to build a user-centric graph model for the detection task.

In spite of numerous approaches, efficient early identification of fake health news still possesses several challenges. 
Detecting misinformation in social media required working with limited data as most users would not even bother reading the news content before sharing them. 
According to a study by Gabielkov et al. \cite{noclicks}, 59\% of the shared URLs are never clicked on Twitter and consequently, these social media users do not read the news article. Therefore, the authors of this work extensively use user profile features and social engagements (posts) information along with statistics of the news recorded in social media during the propagation of news.
Having realized the efficacy of GNNs and analyzed the challenges and limitations in the prominent methods for fake news detection, the authors propose modelling fake news based on social engagements and user specific data along with news and publisher details. 
Social engagements such as tweets and retweets reflect the user's stance on a topic. Furthermore, the credibility of a social media user are attributed by their profile and usage statistics. The historical activity of the account can be utilized to distinguish genuine and bot accounts which further assists in detecting fake news articles. In addition to this, the meta information of news sources serves as a complementary component for fake news detection.

The authors used the HealthStory dataset to develop the proposed model that is capable of identifying health fake news articles.

\section{Dataset}
\label{dataset}
FakeHealth \cite{fakehealth}, is a data repository consisting of two datasets, HealthStory and HealthRelease. HealthStory contains news stories reported by news media such as Reuters Health while HealthRelease consists of official news releases from various sources such as universities, research centres and companies. The repository exclusively comprises of health related articles. HealthStory comprises of 73.6\% of the news articles from FakeHealth implying that only 26.4\% of the remnant articles are present in HealthRelease. Furthermore, the total number of social engagements (tweets, retweets and replies) in HealthStory is 532,380 which is significantly greater than HealthRelease which aggregates to only 65,872. 
This huge difference is attributed to greater number of news articles in the former which is almost  higher than the latter. 
Hence, authors chose to work on HealthStory for the task of fake news detection. 

HealthStory contains a considerably larger set of articles compared to HealthRelease (over 1600 vs around 600) and is therefore used in this work.

HealthStory includes contents of the news article, reviews about the news given by medical experts, engagements and network information for associated users. Twitter mainly consists of the following user engagements -- tweet refers to an original post, retweet is a re-posting of a tweet and reply is a response to a tweet. 

Medical experts rate the news based on ten independent criteria to determine the label (fake/real) of a news article. The news is rated on a 5 point scale and a rating lower than 3 implies that the news is fake. The statistics of the HealthStory dataset are provided in Table \ref{hs summary}.

\begin{table}[]
\renewcommand{\arraystretch}{1.4}
\centering
\caption{HealthStory Summary}
\label{hs summary}
\begin{tabular}{|c|c|c|c|c|c|}
\hline
Tweets  & Retweets & Replies & Articles & True news & Fake news \\ \hline
384,073 & 120,709  & 27601   & 1,690    & 1,218     & 472       \\ \hline
\end{tabular}
\end{table}



\section{Problem Definition}\label{problem defn}
Let $\mathcal{N}$ = \{$n_{1}$, $n_{2}$, $n_{3}$....$n_{\mid \mathcal{N}\mid}$\} denote the set of news articles,  $\mathcal{U}$ denote the set of users who share their views on Twitter. Given a news article $n_{i}$, $\mathcal{T}^{(i)}$ $\subseteq$ $\mathcal{U}$ and $\mathcal{R}^{(i)}$ $\subseteq$ $\mathcal{U}$ be the set of users who post a tweet and retweet a tweet about the article respectively. An engagement is a tweet or retweet made by a user on the news article and is represented by $E$ = [$\omega_{1}$, $\omega_{2}$, $\omega_{3}$,...., $\omega_{e}$] where $\omega$ denotes each word in $E$ and $e$ is the number of words in the engagement. Let $\mathcal{P}^{(i)}$ = [$p_{1}^{(i)}$, $p_{2}^{(i)}$, $p_{3}^{(i)}$....$p_{\mid f \mid}^{(i)}$] denote the news ($n_{i}$) feature vector having ${f}$ features. Each user $u$ $\in$ $\mathcal{U}$ is associated with a $d$-dimensional feature (eg. user profile, historical features) vector $\mathcal{X}$ $\in$ $\mathbb{R}^{1 \times d}$. 
 
 Let $\mathcal{G}_{t}^{(i)}$, $\mathcal{G}_{r}^{(i)}$ be the social interaction graphs for users who engaged on the article. The subscripts denote the engagement (tweet/retweet). Each node on the graph corresponds to a user. The details of the construction of the social interaction graphs are elaborated in Section \ref{socialgraph}. $\mathcal{Y}$ = \{0, 1\} is used to denote the outcome predicted, where y = 0 indicates the news is fake and y = 1 indicates the news is real.

Given set of news article $\mathcal{N}$, $\mathcal{T}$, $\mathcal{R}$ and $\mathcal{P}$, $\mathcal{G}$ = \{$\mathcal{G}_{t}$, $\mathcal{G}_{r}$\} is obtained. This work aims to determine whether the given news article $n_{i}$ is fake or real \eqref{mainidea}.

\begin{equation} \label{mainidea}
    \textbf{F} : \Delta (\mathcal{G}, E, \mathcal{P}) \to \widetilde{\mathcal{Y}}
\end{equation}

This work aims to solve the problem only with the social ($\mathcal{G}$, E) data and the data ($\mathcal{P}$) associated with the news articles.

\section{Architecture}\label{architecture}

The authors propose a novel framework \textbf{SO}cial graph with \textbf{M}ulti-head attention and \textbf{P}ublisher information and news \textbf{S}tatistics \textbf{Net}work (SOMPS-Net) that consists of 2 major components; Social Interaction Graph component (SIG) and Publisher and News Statistics component (PNS). 
\begin{figure}[h]
    \centering
    \includegraphics[width=\linewidth]{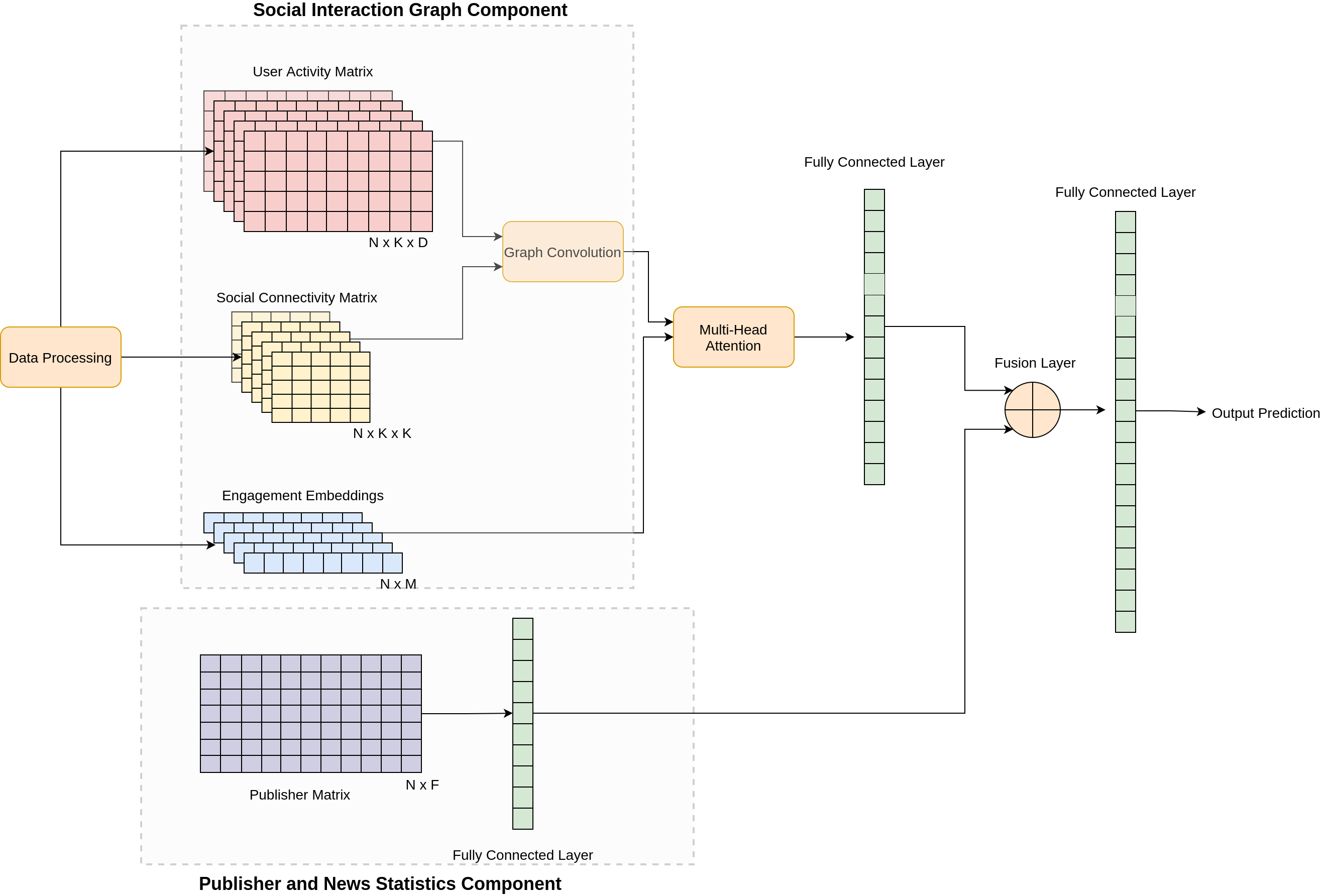}
    \caption{SOMPS-Net}
    \label{paperarchitecture}
\end{figure}

The SIG further consists of 5 sub-components. $(i)$ \textit{Engagement Embedding} to generate a single dense vector representation for all engagements on the article. Separate vectors for tweets and retweets are obtained from this sub-component. $(ii)$ \textit{Social Connectivity Representation} to determine the social connectivity between users based on their followers and following network. $(iii)$ \textit{User Activity Matrix} contains the feature vectors of the users who engage on the article. $(iv)$ \textit{Convolutional} component to obtain the user representation in the social graph. $(v)$ \textit{Cross Attention} component to capture the correlation between the engagement and the users who engaged on the article.

The PNS is introduced to model the data associated with the news article. The truthfulness of the news (fake/real) is predicted by combining the learned representation from SIG and PNS.

\subsection{Social Interaction Graph} \label{socialgraph}
\subsubsection{Engagement Embedding (EE):}
For every news article $n_{i}$ there are $p$ tweets and $q$ retweets posted on Twitter and they have a maximum length of 280 characters. Since each engagement has a variable number of characters, the authors propose a method to handle the inconsistency. Firstly the content of the tweet is tokenized and padded to a maximum sequence length of $m$ which is determined by the median number of words used in all the tweets on the article. The GloVe model \cite{glove} is then used to generate the vector representation of each element in the resulting sequence. The embeddings of all the tweets are then combined and the resultant is a 3-dimensional vector $\prescript{}{t}{}V^{(i)}$ $\in$ $\mathbb{R}^{p \times m \times e}$,  where $e$ is the embedding dimension. Finally, a dense vector  $\prescript{}{t}{}\bar{V}^{(i)}$ $\in$ $\mathbb{R}^{m \times e}$ is obtained by averaging the embedding values across every tweet posted about the article given by Equation \eqref{engagement_embedding}. Similarly, the vector $\prescript{}{r}{}\bar{V}^{(i)}$ is obtained for the retweets made on the article. The averaged engagement embeddings ($\bar{\mathcal{V}}$) obtained from \eqref{engagement_embedding} are then passed through a Bi-Directional LSTM \cite{bilstm} layer. Let $\mathcal{L}$ denote the output from the Bi-LSTM layer.

\begin{equation} \label{engagement_embedding}
\begin{gathered}
\prescript{}{t}{}\bar{V}^{(i)} = \frac{\Sigma^{p}_{j=1} \prescript{}{t}{}V_{j}^{(i)}} {p}, \\
\textrm{where} \; \prescript{}{t}{}V_{j}^{(i)} \textrm{is the word embedded vector representation of a tweet j.}\\
\end{gathered}
\end{equation}

\subsubsection{Social Connectivity Representation (SC):}

For each news $n_{i}$ let $\tau_t^{(i)}$ and $\tau_r^{(i)}$ be the timestamp at which the first tweet and retweet were made about the article. The first $k_{t}$ users who tweet about the article since $\tau_t^{(i)}$ are considered for the experiments. The number of users considered is determined by obtaining the median value given by the formula $k_{t}$ = med\{$\mid \mathcal{T}^{(1)} \mid$, $\mid \mathcal{T}^{(2)} \mid$...$\mid \mathcal{T}^{(\mid \mathcal{N} \mid)} \mid$\}. 

Consider two users $u_{x}$ and $u_{y}$ from the first $k_t$ users. Let $\phi$($u_{x}$) denote the set of users who follow $u_{x}$ and $\psi$($u_{x}$) denote the set of users followed by $u_{x}$. The connectivity score between $u_{x}$ and $u_{y}$ is given by Equation \eqref{connectivity}. As a result, social connectivity matrix $\mathcal{A}_{t}^{(i)}$ $\in$ $\mathbb{R}^{k_{t} \times k_{t}}$ is obtained for the users who tweeted about the news article. Similarly, the social connectivity matrices ($\mathcal{A}_{r}$) for retweet users are obtained. 

\begin{equation}
    Connect(u_{x}, u_{y}) = \frac{\mid (\phi(u_{x}) \cap \phi(u_{y})) \cup (\psi(u_{x}) \cap \psi(u_{y})) \mid}{ \mid \phi(u_{x}) \cup \phi(u_{y}) \cup \psi(u_{x}) \cup \psi(u_{y}) \mid} \label{connectivity}
\end{equation}

\subsubsection{User Activity Matrix (UAM):}Each user who posts about the article is associated with a feature vector ($\mathcal{X}$) as mentioned in Section \ref{problem defn}. In addition to the available profile features, the authors extract features based on historical activity of the account. Features considered are listed in Table \ref{uam features}. As a result, node feature matrices $\mathcal{H}_{t}^{(i)} \in \mathbb{R}^{k_{t} \times d}$ and $\mathcal{H}_{r}^{(i)} \in \mathbb{R}^{k_{r} \times d}$ are generated to represent the node (user) features for a news article ($n_{i}$).

\begin{table}[]
\renewcommand{\arraystretch}{1.3}
\centering
\caption{UAM Features}
\label{uam features}
\resizebox{\columnwidth}{!}{%
\begin{tabular}{|c|c|}
\hline
Whether the account is protected or not & Whether the profile image is default or not \\ \hline
Whether the account is verified or not & Whether the account UI is default or not \\ \hline
Whether geo location is enabled or not&Number of words in the description of the user\\ \hline
Number of words in the username & Number of tweets liked by the user            \\ \hline
Friends count of the user  & Average number of posts made per day            \\ \hline
Followers count of the user            & Maximum number of posts on a single day   \\ \hline

No. public lists the user is member of &  Days between account creation and post          \\ \hline
Number of tweets made on the news article (TW only)& Time between first tweet and retweet in hours (RT only)\\ \hline
\end{tabular}%
}
\end{table}

\subsubsection{Graph Convolution (GC):}

For each $n_{i}$, graph $\mathcal{G}^{(i)}$ is defined by its connectivity matrix $\mathcal{A}^{(i)}$. Two independent graphs ($\mathcal{G}_{t}$, $\mathcal{G}_{r}$) for each engagement (tweet/retweet) are obtained. A graph is $\mathcal{G}$ is represented by a tuple containing set of nodes/vertices and edges/links. The graph can be represented as $\mathcal{G} = (\mathcal{V}, \mathcal{E})$. The nodes of the graph ($\mathcal{V}$) represent the users and the edges ($\mathcal{E}$) represent the connectivity score \eqref{connectivity} between the users. A degree normalized social connectivity matrix  ($\bar{\mathcal{A}}$) is derived using Equation \eqref{normalized matrix} from the degree matrix $\mathcal{D}$. A Graph Convolutional Network (GCN) layer \cite{gcn} is then applied over $\mathcal{G}$ to obtain the graph embeddings. The number of layers in a GCN corresponds to the farthest distance that the node features can propagate. For this work, the authors consider using a 3 layered stacked GCN to capture the finer representation of nodes in the graph network. Let $l$ denote the $l^{th}$ layer of the stacked GCN. The node representations in the $({l+1})^{th}$ layer is given by Equation \eqref{graph conv}. Consequently, the graph representations $\mathcal{G}_{t}$, $\mathcal{G}_{r}$ corresponding to $\mathcal{A}_{t}$, $\mathcal{A}_{r}$ are obtained.

\begin{equation} \label{normalized matrix}
\bar{\mathcal{A}} = \mathcal{D}^{\frac{-1}{2}} \mathcal{A} \mathcal{D}^{\frac{-1}{2}}, 
\end{equation}
\vspace{-15pt}
\begin{gather} \label{graph conv}
    \mathcal{H}^{(l+1)} = \sigma(\bar{\mathcal{A}}\mathcal{H}^{(l)}\mathcal{W}^{(l)})
\end{gather}
where, \\
\hspace*{3em}
\begin{tabular}{rl}
    $W^{(l)}$ & is the weight matrix of the $i^{th}$ graph convolutional layer\\
    $\mathcal{D}_{ii}$ = & $\Sigma_{j}\mathcal{A}_{ij}$ \\
    $\sigma$  & is the activation function\\
\end{tabular}

\subsubsection{Multi Head Attention (MHA):}

Attention mechanism can be described as the weighted average of (sequence) elements with weights dynamically computed based on an input query and element's key. Query (Q) corresponds to the sequence for which attention is paid. Key (K) is the vector used to identify the elements that require more attention based on Q. The attention weights are averaged to obtain the value vector (V). A score function \eqref{attention} is used to determine the elements which require more attention. The score function takes Q and K as input and outputs the attention weight of the query-key pair. In this work, the authors consider using the scaled dot product proposed by Vaswani et al. \cite{attentionisallyouneed}. The attention weights are calculated based on the graph embeddings ($\mathcal{G}$) and the Bi-LSTM output ($\mathcal{L}$). K and V are initialized with the value of $\mathcal{G}$ and Q is initialized with the value of $\mathcal{L}$. 

The scaled dot product attention captures the characteristics of the sequence
it attends. However, often there are multiple different aspects to a sequence, and
these characteristics cannot be captured by a single weighted average vector.
The Multi-Head Attention (MHA) \cite{attentionisallyouneed} uses multiple different query-key-value
triplets (heads) on the same features. The Q, K and V matrices are transformed
into sub-queries, sub-keys and sub-values and are then passed through the scaled
dot product \eqref{attention} attention independently. The attention outputs from each head
are then combined and the final weight matrix (WO)is calculated and dk is
the hidden dimensionality of K. Thus the output OS from this component is
obtained

\begin{equation} \label{attention}
Attention (Q, K, V) = Softmax(\frac{QK^{T}}{\sqrt{d_{k}}})V 
\end{equation}

\begin{equation} \label{multiheadattention}
\begin{gathered}
MultiHead(Q, K, V) = Concat(head_1, \dots, head_n)W^O  \\
\textrm{where} \; \textrm{$head_i$} = Attention(QW^Q_i, KW^K_i, VW^V_i), \\
\textrm{$W^{Q}$, $W^{K}$, $W^{V}$ are the weight matrices of Q, K and V respectively}
\end{gathered}
\end{equation}

\subsection{Publisher and News Statistics}\label{pubandn}
The Publisher and News Statistics (PNS) is the second component in the SOMPS-Net framework. The intention of the fake news spreader is to proliferate the news instantaneously and provoke chaos amongst the targeted audience. Thus, the authors propose that utilizing the statistical information of the news, recorded throughout its lifetime along with the metadata of the news article could help in the detection task. Further, credibility information about a news publisher is another significant factor that helps in determining the authenticity of the news. The features considered are listed in Table \ref{pn features}. The feature vector ($\mathcal{P}$) is passed through a dense layer and the output $O^{P}$ from this component is obtained.

\begin{table}[h]
\renewcommand{\arraystretch}{1.5}
\centering
\caption{PNS Features}
\label{pn features}
\begin{tabular}{|c|c|}
\hline
Total number of tweets           & Number of unique users mentioned \\ \hline
Total number of retweets         & Lifetime of the news in days           \\ \hline
Total number of replies          & Tags associated with article           \\ \hline
Total number of unique hashtags  & News publisher                         \\ \hline
Total number of likes (any engagement) & Average rating of the news publisher \\ \hline
\end{tabular}
\end{table}


The outputs $O^{S}$ and $O^{P}$ are fused together and the resulting high dimensional vector is passed through a fully connected layer. The final outcome $\mathcal{Y}^{(i)}$ for $n_{i}$ is predicted by the model as illustrated in Equation \eqref{concatcomp}.

\begin{equation} \label{concatcomp}
	\Gamma (O^{S} \oplus O^{P}) \to Y \\
\end{equation}

\section{Experiments and Results}\label{results}

\subsection{Experimental Setup} \label{exptsetup}


The performance of SOMPS-Net is evaluated and compared based on Accuracy and F1-score. The news articles were proportionally sampled (stratified) and the data was split into 75\% for training, 10\% for validation and 15\% for testing. The authors consider news articles that have at least one of each engagement (tweet/retweet). As a result, 1492 news articles ($\mathcal{N}$) were obtained. The number of real and fake news articles obtained were 1082 and 410 respectively. .

The following are the hyperparameter settings of used in SGM: number of tweet users ($k_t$) : 32, number of retweet users ($k_r$) : 16, engagement (tweet/retweet) length ($m$) : 20, number of news features ($f$) : 10, number of user associated features ($d$) : 13 (tweet), 14 (retweet), word embedding dimension ($e$) : 100, number of GCN layers ($l$) : 3, GCN output dimension : 16, number of hidden units in Bi-LSTM : 100, number of attention heads ($n$) : 16, size of each attention head for key (K) and query (Q) : 4, size of attention head for value (V) : 12, learning rate : 0.001, dropout : 0.5, optimizer : SGD (Gradient descent with momentum), loss function : binary cross entropy.



\subsection{Comparison systems} \label{comparesystem}
We compare the performance of the proposed architecture with the previous works done on HealthStory.

\subsubsection{Dai et al.} \cite{fakehealth} considered three methods -- ($i$) linguistic-based, ($ii$) content-based and ($iii$) social context-based for fake news detection. In linguistic based methods, the authors used Logistic Regression, SVM and Random Forest for modelling the lexicon-level features. CNN and Bi-directional GRU were employed for content-based modelling. Finally, in social context based methods, the authors used the Social Article Fusion (SAF) model initially proposed by \cite{saf}. The SAF model utilizes user embeddings and replies. The social context features learned from an LSTM encoder were combined with the former to make the final prediction.


\subsubsection{Chandra et al.} \cite{SAFER} proposed a framework SAFER which uses graph-based model for fake news detection. The framework aggregates information from the content of the article, content sharing behaviour of users and the social connections of the users. SAFER consists of two components -- graph encoder and text encoder component. The graph encoder takes the community graph of the users and the text encoder takes the text of the article as inputs. The outputs from the two components are then concatenated and passed through a logistic classifier. The authors considered six different GNN architectures for generating user embeddings.

\subsubsection{SOMPS-Net (This work)} The results using the proposed novel framework SOMPS-Net are compared with other systems for its robustness. To validate the importance of each of the components in the proposed framework, the authors also consider 2 variants of SOMPS-Net -- $SOMPS_{SIG}$ and $SOMPS_{PNS}$. In $SOMPS_{SIG}$ only the SIG component is considered and in $SOMPS_{PNS}$ only the PNS component is considered. Equation \eqref{modelVariants} represents the components and the data used in the 3 variants.

\begin{equation}\label{modelVariants}
\begin{split}
   F_{SOMPS} : \Delta (\mathcal{G}, E, \mathcal{P}) \to \widetilde{\mathcal{Y}} \\
    F_{SOMPS_{SIG}} : \Delta (\mathcal{G}, E) \to \widetilde{\mathcal{Y}}\\
    F_{SOMPS_{PNS}} : \Delta (\mathcal{P}) \to \widetilde{\mathcal{Y}}
\end{split}
\end{equation}
\subsection{Results analysis}
Table \ref{mainresults} illustrates the results obtained using SOMPS-Net and other comparison systems considered.

The SOMPS-Net framework outperforms linguistic-based and content-based models used in Dai et al. \cite{fakehealth} by 6.1\% and 6.6\% respectively. This further solidifies the initial hypothesis of considering the social engagement data to model fake news detection. On comparison, social context-based model used in \cite{fakehealth} has around 4\% accuracy improvement over SOMPS-Net. However, SOMPS-Net performs better than this model by 4\% when F1 score is considered. Moreover, the social context model uses replies made on the article. Upon exploratory analysis it was found that only 720 articles contained replies. Thus an accuracy of 76\% was achieved using only 541 true and 179 fake news articles.
On the other hand, the authors of this work considered 1492 articles as mentioned in Section \ref{comparesystem}. Since, the social context model considered by \cite{saf} uses less than 50\% of the HealthStory articles, it has limited applicability. 



SAFER achieved an F1-score of 62.5 \%, and SOMPS-Net outperformed SAFER with a relative 27.36\% performance improvement, asserting its superior performance. Also, it can be inferred from Table \ref{mainresults} that SOMPS-Net outperforms each of the six different GNN architectures considered by SAFER. Furthermore, the robustness of the SOMPS-Net's components are well established since each of the components in SOMPS-Net -- $SOMPS_{SIG}$ and $SOMPS_{PNS}$ outperformed SAFER by 16\% and 12.2\% respectively. Additionally, it is also observed that SAFER uses the news content for modelling while SOMPS-Net uses only the social context and metadata information about the news. This proves the initial hypothesis of considering only the social context and publisher information for detecting fake news articles since SOMPS-Net significantly outperformed SAFER.

\begin{center}
    
\begin{threeparttable}[h]
\renewcommand{\arraystretch}{1.2}
\caption{Fake News Detection Results}
\label{mainresults}
\begin{tabular}{|c|c|c|c|}
\hline
\textbf{Model} & \textbf{Approach}       & \textbf{Accuracy} & \textbf{F1 Score} \\ \hline
Dai et al.  & Linguistic-based        & 0.720             & 0.735       \\ \hline
Dai et al.  & Content-based           & 0.742             & 0.730       \\ \hline
*Dai et al. & Social context-based    & 0.760             & 0.756       \\ \hline
SAFER & Graph + Content  & Not specified     & 0.625       \\ \hline
SOMPS- P        & Publisher \& News             & 0.727             & 0.747       \\ \hline
SOMPS- SIG      & Graph                   & 0.727             & 0.785       \\ \hline
\textbf{SOMPS-Net}            & \textbf{Graph + Publisher \& News}       & \textbf{0.727}             & \textbf{0.796}       \\ \hline
\end{tabular}
\begin{tablenotes}
      \small
      \item ${}^\star$ Uses only 720 (42.6\%) articles
    \end{tablenotes}
\end{threeparttable}
\end{center}

\subsection{Early Detection}
\begin{figure} 
    \centering
    \includegraphics[width=0.8\columnwidth]{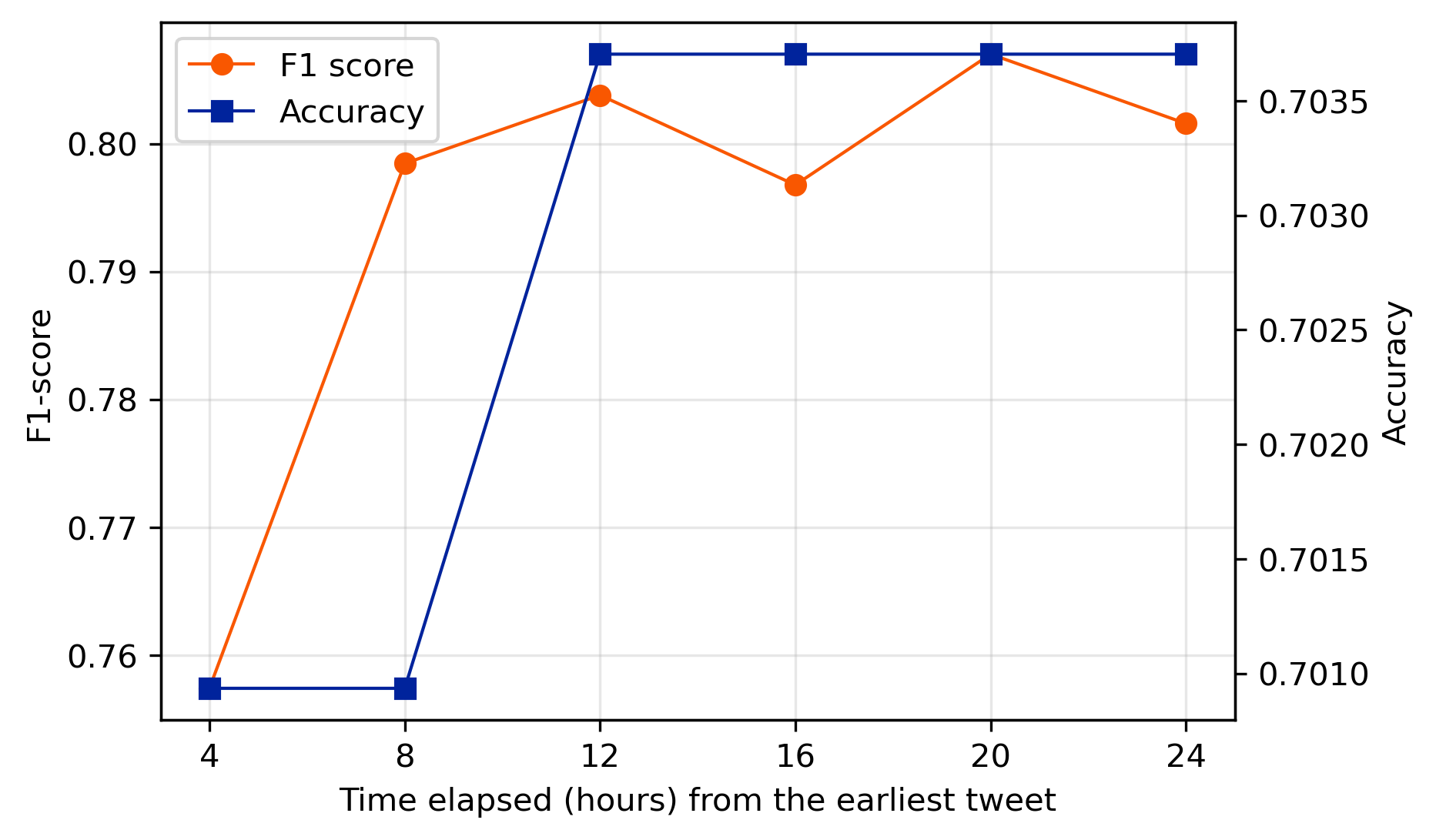}
    \caption{Early Detection Results}
    \label{edresults}
\end{figure}

Early detection of fake news is crucial to restrain its reach from wider audience, particularly for health related information. The task of early detection is driven by social engagements of the news article captured within a fixed time frame. Time intervals in multiples of 4 since the first tweet about the article were considered.
The performance of SOMPS-Net framework for each time interval is illustrated in Figure \ref{edresults}.

From Fig. \ref{edresults} the authors infer the following. The detection rate of the model in terms of F1-score improves drastically till 8 hours and sustains around 80\%. Maximum F1-Score of 0.807 was achieved from 20 hours of its broadcast. It can be noted that the model detects with an appreciable score of 75\% within just 4 hours which is higher than the comparison systems mentioned in Section \ref{comparesystem}. Hence, it is evident that SOMPS-Net is robust and is capable of detecting fake health articles at its early stages with minimal information. This analysis could therefore help control further diffusion of fake news in social media. 

\section{Conclusion}\label{conclusion}


In this work, the authors propose and successfully test a novel graph based framework -- SOMPS-Net on the HealthStory dataset to detect fake news in the health domain. The proposed framework utilizes the social context data and social reach data of the news article. It consists of 2 components -- Social Interaction Graph (SIG) which consists of 5 sub-components (Section \ref{socialgraph}) utilizing the social context data such as tweets, retweets and user profile features. The authors use a rich user profile feature set (as illustrated in Table \ref{uam features}) that contains user metadata and features extracted from the historical activity of the account. 
The Publisher and News Statistics (PNS) component (Section \ref{pubandn}) utilizes the metadata and statistics of the news article (illustrated in Table \ref{pn features}).

SOMPS-Net performs significantly better than other well-established graph based approaches on HealthStory. A 27.36\% relative performance improvement is achieved from the state-of-the-art graph based models. The importance of each of the components is established since the components -- SIG and PNS outperform SAFER by 16\% and 12.2\% respectively. The authors further exhibit SOMPS-Net's robustness by using data captured within a certain time frame. The model detected fake news with 79.8\% certainty with only 8 hours of data and achieved a maximum F1-score of 80.7\% with 20 hours of information.

For future work, the authors aim to include other modalities into the proposed framework. Also, there exists a need for interpretable and explainable machine learning solutions for health news articles. Providing deep insights about the model and an explanation on why the news is classified as fake or real can be critical in mitigating fake news spread and also safeguard the consumers from adversaries. 

\bibliographystyle{splncs04}
\bibliography{reference}

\end{document}